%% file: main.tex
\documentclass{article}
\usepackage{spconf,amsmath,graphicx}
\usepackage{hyperref}
\usepackage{url}
\usepackage{graphicx}
\usepackage{subcaption}

\graphicspath{{figures/}}
\input{macrosjvg}

\title{Semi-supervised lane detection with Deep Hough Transform}

\name{Yancong Lin \qquad Silvia-Laura Pintea \qquad Jan van Gemert}
\address{Vision Lab, Delft University of Technology, The Netherlands}

\begin{document}

\maketitle

\begin{abstract}

Current work on lane detection relies on large manually annotated datasets. We reduce the dependency on annotations by leveraging massive cheaply available unlabelled data. 
We propose a novel loss function exploiting geometric knowledge of lanes in Hough space, where a lane can be identified as a local maximum. By splitting lanes into separate channels, we can localize each lane via simple global max-pooling. The location of the maximum encodes the layout of a lane, while the intensity indicates the the probability of a lane being present. Maximizing the log-probability of the maximal bins helps neural networks find lanes without labels. On the CULane and TuSimple datasets, we show that the proposed Hough Transform loss improves performance significantly by learning from large amounts of unlabelled images.

\end{abstract}

\input{sections/01_introduction}

\input{sections/02_related_work}

\input{sections/03_method}
\input{sections/04_experiments}
\input{sections/05_discussion}

{\small
\bibliographystyle{IEEEbib}
\bibliography{ref}
}
\end{document}

%% file: macrosjvg.tex
\usepackage{comment} 

\usepackage{xcolor}

\newcommand{\fig}[1]{Fig.~\ref{fig:#1}}
\newcommand{\tab}[1]{Table~\ref{tab:#1}}

\usepackage{pifont}
\usepackage{caption}
\usepackage{tabularx} 
\newcolumntype{Y}{>{\centering\arraybackslash}X}

\usepackage{booktabs}
\usepackage{multirow}

%% file: sections/01_introduction.tex
\section{Introduction}
\label{sec:intro}

One key component of self-driving cars is the lane-keeping assist \cite{he2016accurate,pohl2007driver}, which actively keeps the vehicles in the marked lanes. The lane-keeping assist relies on accurate lane detection in the wild, which is a highly challenging task because of illumination and appearance variations, traffic flow, and new unseen driving scenarios \cite{hou2019learning}.

State-of-the-art deep learning methods for lane detection perform remarkably well on benchmark datasets \cite{hou2019learning, hou2020inter, romera2017erfnet, qin2020ultra}. 
However, they rely on deep networks powered by massive amounts of labelled data. 
Although the data itself can be obtained at relatively low cost, it's their annotations that are laborious and thus expensive~\cite{huijser2017active}.
Moreover, the existing curated datasets do not cover all the possible driving scenarios that could be encountered in real-world situations.
Being able to leverage additional realistic unlabelled training data would allow for a more robust lane detection system.

To make effective use of additional unlabelled data, we propose a semi-supervised Hough Transform-based loss which exploits geometric prior knowledge of lanes in the Hough space \cite{duda1972use, lin2020deep}. 

Lanes are lines, thus we propose a semi-supervised Hough Transform loss that parameterizes lines in Hough space, by mapping them to individual bins represented by an offset and an angle. 
Inspired by the work in \cite{lin2020deep}, we rely on a trainable Hough Transform and Inverse Hough Transform ($HT\text{-} IHT$) module embedded into a neural network to learn Hough representations for lane detection.
We subsequently extend its use for semi-supervised training, by noting that the presence of lanes leads to Hough bins with maximal votes. 
Maximizing the log-probability of these Hough bins requires no human supervision,  enabling the network to detect lanes in unlabelled images.

This paper makes the following contributions: $(1)$ we present an annotation-efficient approach for lane detection in a semi-supervised way; $(2)$ to this end, we propose a novel loss function to exploit prior geometric knowledge of lanes in Hough space; $(3)$ we experimentally show improved performance on the CULane \cite{pan2017spatial} and TuSimple \cite{tusimple} datasets, given large amounts of unlabelled data.

%% file: sections/02_related_work.tex
\section{Related Work}

\noindent\textbf{Lane detection methods.}
Classic work on lane detection is based on knowledge-based manually designed geometric features. Examples include grouping image gradients ~\cite{kim2008robust,  deusch2012random, yoo2013gradient, wu2012practical}, or 
line detection techniques through Hough Transform \cite{dorj2016precise,ghazali2012road,wang2000lane,yu1997lane} relying on local edges extracted using image gradients. 
A main drawback of such knowledge-based methods is their inability to handle complex scenarios where traffic flow and illumination conditions change dramatically. 
Here, we address this by learning appearance variation of lanes in a deep network, while still relying on the Hough Transform as prior knowledge for line detection~\cite{duda1972use, lin2020deep}.

Recently, deep neural networks have been employed for efficient lane detection, replacing well-engineered features. Typically, the learning-based methods treat the lane detection as a semantic segmentation task and learn semantic features from large datasets \cite{he2016accurate,lee2017vpgnet,li2016deep,neven2018towards,tang2020review,zhang2018geometric}.
In contrast to these works we improve the prediction accuracy by leveraging massive unlabelled data through semi-supervised learning.

\smallskip\noindent\textbf{Semi-supervised methods.}
Semi-supervised methods solve the learning task by relying on both labelled and unlabelled data \cite{van2020survey}, and are divided into: inductive approaches constructing a classifier over labelled and unlabelled data \cite{blum2004semi,chen2020simple,kingma2014semi}, and transductive approaches propagating where the task information is shared between data points \cite{blum2001learning,jebara2009graph,zhu2003semi}.
A self-driving car has no access to the test statistics, therefore  we consider the inductive case.

%% file: sections/03_method.tex
\section{Semi-supervised lane detection}
\begin{figure}[t!]
    \centering
    \hspace*{-10px}    
    \includegraphics[width=0.5\textwidth]{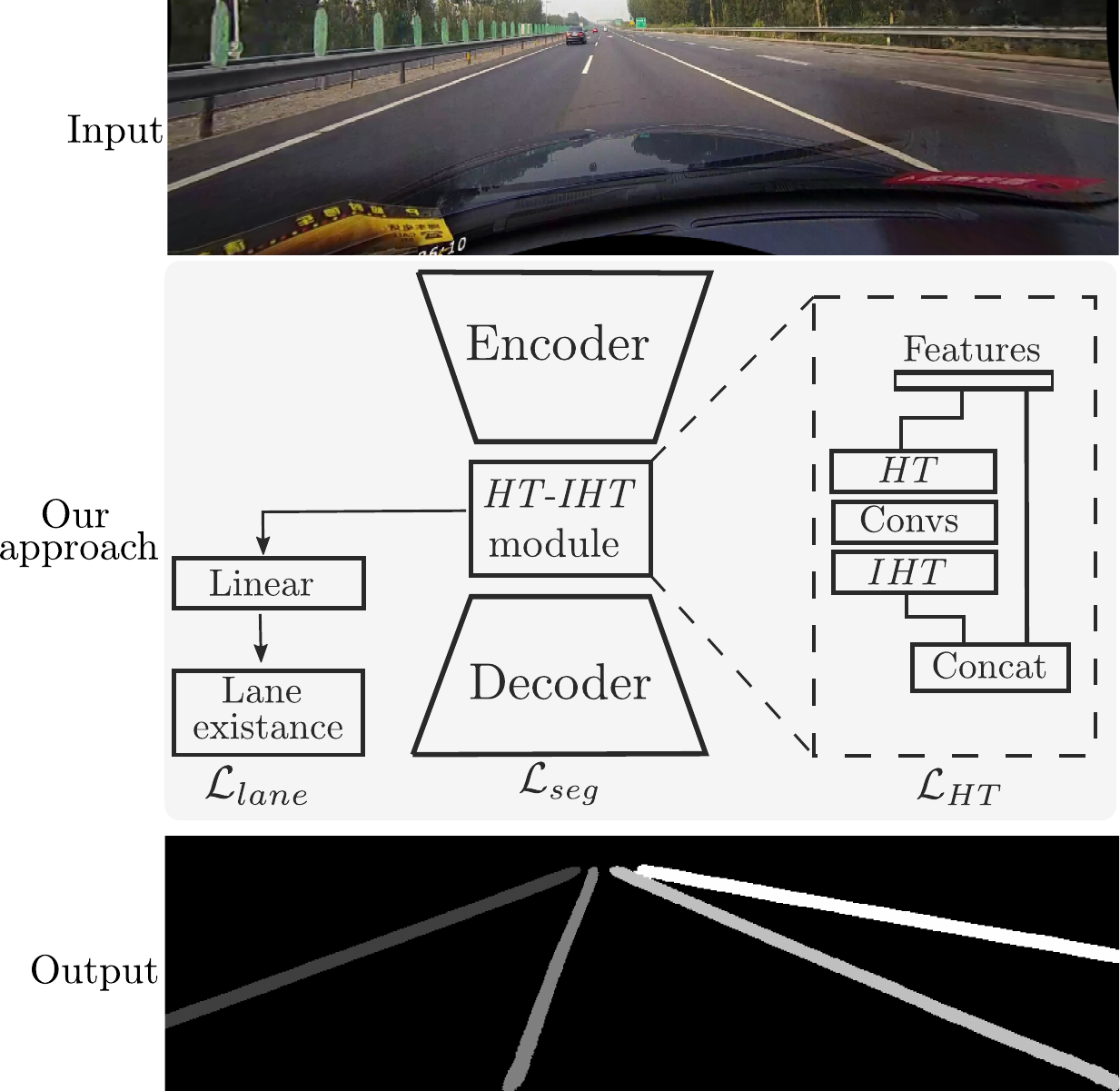}
    \caption{\textbf{Overview of our model.} 
    We have an encoder, a decoder, and a fully connected layer inspired by the ERFNet \cite{hou2020inter, romera2017erfnet} with a trainable Hough-Transform (HT) and Inverse Hough-Transform (IHT)  module \cite{lin2020deep}, on top of which we build our  $HT$-based semi-supervised loss maximizing the probability of the maximal bins in Hough space, where $\mathcal{L}_{lane}$, $\mathcal{L}_{seg}$, and $\mathcal{L}_{HT}$ are the optimized loss functions. 
    }
    \label{fig:model}
\end{figure}

Given an input image, our model outputs a lane probability and a semantic segmentation mask of lane pixels. 
We use as a starting point the popular ERFNet \cite{romera2017erfnet}\footnote{We rely on the implementation in \cite{hou2020inter}: \url{https://github.com/cardwing/Codes-for-Lane-Detection}}.
The ERFNet contains a convolutional encoder for deep feature extraction, a convolutional decoder for lane predictions, and a fully connected layer for predicting the probability of a lane. 
We insert a trainable Hough Transform and Inverse Hough Transform (\emph{HT-IHT}) block \cite{lin2020deep} between the encoder and decoder, and utilize the Hough representations of lanes for semi-supervised learning.
\fig{model} depicts the overall structure of our model. 
\subsection{Hough Transform line priors}

We encode an input image to a semantic feature representations $F$ which is mapped to the Hough space, through a trainable Hough Transform module \cite{lin2020deep}. 
The Hough transform $HT$ maps a feature map $F$ of size $[H \times W]$ to an $[N_{\rho} \times N_{\theta}]$ Hough histogram, where $N_{\rho}$ and $N_{\theta}$ are the number of discrete offsets and angles.
Pixels along lines in $F$ are mapped into discrete pairs of offsets $\rho$ and angles $\theta$.
Specifically, given a line direction indexed by $i$ with its corresponding pixels $(x_{i}, y_{i})$, they all vote in the Hough space for the closest bin ($\rho$, $\theta$): 
\begin{align}
    HT(\rho, \theta) = \sum_i F(x_{i},  y_{i}), 
\end{align}
where the mapping is given by $\rho = x_{i} \cos \theta + y_{i} \sin \theta$.

We perform a set of 1D convolutions in Hough space over the offset direction and apply an Inverse Hough Transform $IHT$ module  mapping the $[N_{\rho} \times N_{\theta}]$ Hough histogram back to an $[H \times W]$ feature map~\cite{lin2020deep}. 
The $IHT$ maps bins ($\rho$, $\theta$) to pixels $(x_{i}, y_{i})$ by averaging all the $HT$ bins where a certain pixel has voted: 
\begin{alignat}{1}
    IHT(x_{i},y_{i}) &= \frac{1}{N_\theta}\sum_\theta HT(x_{i} \cos \theta + y_{i} \sin \theta, \theta).
    \label{eq:iht}
\end{alignat} 
We concatenate the features $F$ with the $IHT$ features, followed by a convolutional layer merging these two branches. 
We set $H=26$, $W=122$, $N_{\rho}=125$ and $N_{\theta}=60$.  

\subsection{Hough Transform loss for unlabelled data}
A lane is composed of a set of line segments with a certain width, that share the same orientation. 
For unlabelled images we rely on the observation that lanes correspond to local maxima in the Hough space. Since the ERFNet \cite{hou2020inter, romera2017erfnet} predicts a single lane in each output channel, the mapping to Hough space recovers the lanes as global maxima in their respective channels. Having a large global maximum indicates that pixels along that line direction are well aligned, thus falling in the same bin.
Based on this observation, we provide supervision to unlabelled inputs by maximizing the log-probability of the maximum bin $(\hat{\rho}, \hat{\theta})$ in Hough domain. 
To give the $HT$ bins a probabilistic interpretation, we rescale the $HT$ maps between $[0,1]$ for each angle direction independently by applying an $L_1$ normalization over the offset dimension:
\begin{align}
    \mathcal{L}_{HT} = -\log \left(  \frac{HT(\hat{\rho}, \hat{\theta})}{ \sum_{k=0}^{N_{\rho}} HT(\rho_k, \hat{\theta})} \right), 
     \label{eq:l_ht}
\end{align}
where $(\hat{\rho}, \hat{\theta})$ is the positions of the global maximum in Hough space, calculated from the predicted segmentation masks.



\subsection{Training with both labelled and unlabelled data}
We train our model with both labelled and unlabelled data. 
As in \cite{hou2019learning,hou2020inter} the network predicts for each channel a mask used in a cross entropy loss $\mathcal{L}_{\text{seg}}$ over labelled data for predicting the semantic segmentation.
Additionally, the network predicts lane probabilities $p$ which are used in a binary cross entropy loss over labelled data $\mathcal{L}_{\text{lane}}$ for optimizing for the existence of a lane. 
We also optimize the proposed $\mathcal{L}_{HT}$, only when the predicted probability $p$ of a lane is larger than a threshold $\tau$; otherwise, we skip the corresponding lane. 
We set $\tau=0.9$. The total loss is the combination of the three losses: 
\begin{equation}
    \mathcal{L}_{\text{total}} = \mathcal{L}_{\text{seg}} + \alpha \mathcal{L}_{\text{lane}} + \beta \mathcal{L}_{HT}(p, \tau),
    \label{eq:l_total}
\end{equation}
where $\alpha$ and $\beta$ are used to balance different loss terms.


%% file: sections/04_experiments.tex
\section{Experimental analysis}
\noindent \textbf{Datasets.}  
We evaluate our models on the TuSimple dataset \cite{tusimple} and CULane dataset \cite{pan2017spatial}. 
All video clips in TuSimple dataset are taken on highways.
There are 3,626 frames for training and 2,782 frames for testing. 
The CULane dataset contains images from 9 different driving scenarios, such as lanes in shadow and at night with poor lighting conditions. 
There are 88,880 images for training, 9,675 for validation, and 34,680 images for testing. 
We follow the official evaluation protocol to measure accuracy on the TuSimple, and use $F_{1}$ measure on the CULane dataset.

\smallskip\noindent \textbf{Baselines.} 
We compare with the baseline ERFNet \cite{romera2017erfnet}, and with the ERFNet-HT using the \emph{HT-IHT} block \cite{lin2020deep}. 
Both models are trained from scratch with labelled data only. 
For semi-supervised learning, we consider the ERFNet-$pseudo$ pseudo-labeling baseline, and our proposed ERFNet-HT-$\mathcal{L}_{HT}$. 
The ERFNet-$pseudo$ baseline first learns to predict lanes on annotated data only, and subsequently uses the predicted pseudo-labels to annotate unlabelled data, and then retrains the model on all data. 
ERFNet-$pseudo$ treats the prediction with a confidence score larger than $0.9$ as "ground truth" and optimize the $\mathcal{L}_{\text{seg}}$ with pseudo-labels.
ERFNet-HT-$\mathcal{L}_{HT}$ uses our proposed $\mathcal{L}_{HT}$ loss. 
ERFNet-HT-$pseudo$+$\mathcal{L}_{HT}$ combines both pseudo-labelling and our proposed $\mathcal{L}_{HT}$ loss. Additionally, we also compare with s4GAN \cite{mittal2019semi}, a state-of-the-art semi-supervised learning model for semantic segmentation. 

\smallskip\noindent \textbf{Implementation details.} 
We follow the implementation and hyper-parameters in \cite{hou2020inter}. 
We use SGD \cite{bottou2010large} to train ERFNet and ERFNet-HT for 24 epochs. 
ERFNet-$pseudo$,  ERFNet-HT-$\mathcal{L}_{HT}$  and ERFNet-HT-$pseudo$+$\mathcal{L}_{HT}$ are trained with extra unlabelled data for another 12 epochs. 
The initial learning rate is $1 \times 10^{-2}$, and is decreased by a factor of $(1-t/T)^{0.9}$, where $t$ is the current training epoch and $T$ is the total number of epochs, as in \cite{hou2020inter}. 
The batch size is set to be 16. 
For our $\mathcal{L}_{\text{total}}$, we set the weights $\alpha=0.1$ and $\beta=0.01$ to ensure that all loss terms have similar magnitudes. 
Following \cite{hou2020inter}, we multiply the $\mathcal{L}_{\text{seg}}$ for the background class by $0.4$ to counter the large number of background pixels. For s4GAN \cite{mittal2019semi}, we directly use the official implementation \footnote{https://github.com/sud0301/semisup-semseg}.

\begin{table}[t]
    \centering
    \caption{
   \textbf{Performance on TuSimple and CULane datasets with various amounts of labelled and unlabelled data.} The first column indicates the proportion of labelled data for training. The remaining data is treated as unlabelled for semi-supervised learning. 
 ERFNet-HT-$\mathcal{L}_{HT}$ and    ERFNet-HT-$pseudo$+$\mathcal{L}_{HT}$ show performance improvements on both datasets. 
    When the number of labelled samples decreases, the advantage of ERFNet-HT-$\mathcal{L}_{HT}$ is more pronounced. 
    }
    \label{tab:result}
    \renewcommand{\arraystretch}{0.9}
    \resizebox{1\linewidth}{!}{
    \setlength{\tabcolsep}{1mm}{
    \begin{tabularx}{\linewidth}{lYYYYYYY}
        \toprule
        \textbf{Labels} & s4GAN \cite{mittal2019semi} & \multicolumn{5}{c}{ERFNet models} \\ \cmidrule{3-7}
        & & Baseline \cite{romera2017erfnet} & HT\break \cite{lin2020deep} & $pseudo$ & HT-\small{$\mathcal{L}_{HT}$}  & HT-\small{$pseudo$} + {$\mathcal{L}_{HT}$}\\
        \cmidrule{2-7}
        & \multicolumn{6}{c}{\textbf{Accuracy (\%) on the TuSimple dataset}}\\ 
        \midrule
       \textbf{100\%}   & - &\textbf{93.71} & \textbf{93.71} & - & - & - \\
       \textbf{50\%}    & 88.82 & 92.97 & 93.47 &  93.37 & 93.63 & \textbf{93.70} \\
       \textbf{10\%}    & 86.25 & 82.97 & 77.71 & 92.12  & 92.98& \textbf{93.05} \\
        \midrule
        & \multicolumn{6}{c}{\textbf{$F_{1}$ scores on the CULane dataset}}\\
        \midrule
        \textbf{100\%}   & - & 69.86   & \textbf{70.52} & -   & - & - \\
       \textbf{50\%}     & - & 69.39   & 68.59  & 69.68 & \textbf{70.75} & 70.41 \\
       \textbf{10\%}     & - & 60.99   & 61.46   & 65.56 & 64.04 &  \textbf{66.10}  \\
        \textbf{5\%}     & - &  56.61  & 57.78   &  61.99  & 62.32 & \textbf{63.67} \\
        \textbf{1\%}     & - & 32.99   & 32.48   & 51.38 & \textbf{55.10} & 52.80 \\
        \bottomrule
    \end{tabularx}
    }}
\end{table}

\begin{figure*}[t]
    \centering
    \includegraphics[width=1\textwidth]{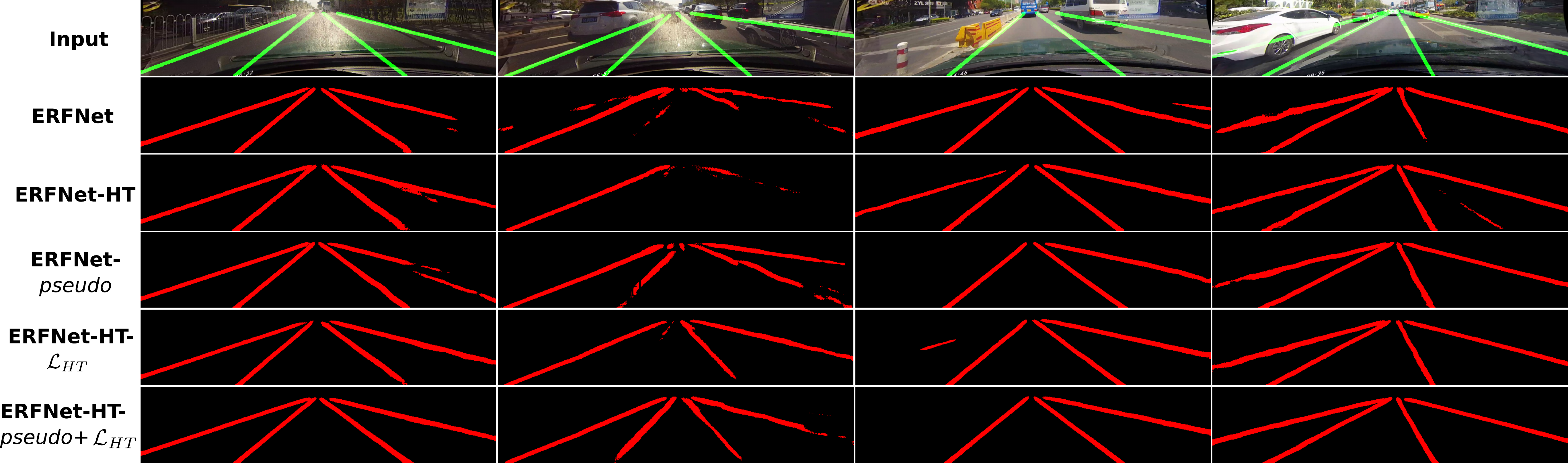}
    \caption{\textbf{Visualizations of predicted lanes on the CULane dataset.} Only $10 \%$ annotated data is used for training. ERFNet-HT-$pseudo$+$\mathcal{L}_{HT}$ performs better on challenging samples and better localizes lane boundaries. The inference speed of the ERFNet-HT is around 13 frames per second on a NVIDIA GTX1080Ti GPU.
    }
    \label{fig:vis}
\end{figure*}

\begin{table}[t]
    \centering
    \caption{\textbf{$F_1$ scores for different scenarios, with $1\%$ labelled data.} 
    ERFNet-HT-$\mathcal{L}_{HT}$ outperforms other models in most scenarios, indicating that the $\mathcal{L}_{HT}$ loss exploits useful geometric knowledge of lanes when adding unlabelled samples.
    }
    \label{tab:culane_1}
    \renewcommand{\arraystretch}{0.9}
    \resizebox{1\linewidth}{!}{
    \begin{tabularx}{\linewidth}{lYYYYY}
    \toprule
    ERFNet models & Baseline \cite{romera2017erfnet} & HT \cite{lin2020deep} &
    \scriptsize{$pseudo$} &
    HT-\scriptsize{$\mathcal{L}_{HT}$}  &
    HT-\scriptsize{$pseudo$} +{$\mathcal{L}_{HT}$}
    \\ 
    \midrule
    Normal~& 49.24 & 51.25& 69.72 & \textbf{75.06} & 71.83\\
    Crowded~  & 31.74 & 31.49 &49.53 & \textbf{52.52} & 50.97\\
    Night~ & 22.36 & 21.67 & 45.27 & \textbf{50.77} & 45.42\\
    No line~ & 18.78 & 17.54 & 28.05 & \textbf{32.02} & 30.12\\
    Shadow~ & 24.71 & 17.69 & 36.63 & \textbf{38.50} & 35.97\\
    Arrow~ & 39.39 & 38.22 & 57.28 & \textbf{63.33} & 59.18\\
    Dazzle~ & 26.25 & 23.76 & \textbf{40.29} & 40.28 & 39.42\\ 
    Curve~ & 33.62 & 34.53 & 46.56 & \textbf{50.52} & 46.42\\
    Cross~\footnotemark~ & 6949 & 8711 & \textbf{3355}& 5292 & 3676\\ 
    \midrule
    Avg $F_1$ & 33.00 & 32.48 & 51.38 & \textbf{55.10} & 52.80\\ 
    \bottomrule
    \end{tabularx}
    }
\end{table}
\footnotetext{For cross-road, we show only the number of false-positives, as in \cite{pan2017spatial}.}

\smallskip\noindent \textbf{Results analysis.}
To evaluate the effectiveness of our $\mathcal{L}_{HT}$ in utilizing unlabelled data, we randomly split the CULane training data into $\{100/0, 50/50, 10/90, 5/95, 1/99 \}$ sets, where the first digit indicates the proportion of labelled data, while the second one is the proportion of unlabelled data. The TuSimple dataset is split into $\{100/0, 50/50, 10/90 \}$ sets, as it contains only 3,626 images.
We use the same splits for all models. We report accuracy on TuSimple and $F_{1}$-measure on the CULane dataset.

\tab{result} compares all models on various training sets. 
ERFNet-HT-$pseudo$+$\mathcal{L}_{HT}$ achieves the best performance on both $50\%$ and $10\%$ subsets of TuSimple dataset. 
The improvement over the supervised baseline is more than $15\%$ on the $10\%$ subset. All semi-supervised ERFNet models improve accuracy, indicating the potential of exploiting massive unlabelled data. 
Pseudo-labeling allows learning from high confidence predictions explicitly, while $\mathcal{L}_{HT}$ optimizes line feature representations in Hough space in an implicit way. However, s4GAN \cite{mittal2019semi} shows inferior performance to other models, due to the fact that s4GAN is not specifically optimized for lane detection, where image content differs substantially from its origin usage.  In general, semi-supervised models perform similar on the TuSimple dataset as it only includes the highway scenario. 
On the CULane dataset, ERFNet-HT-$pseudo$+$\mathcal{L}_{HT}$ consistently outperforms ERFNet-$pseudo$, validating the usefulness of the Hough priors ($\mathcal{L}_{HT}$) in exploiting lane representations in the semi-supervised setting. The s4GAN is lacking since we are unable to produce reliable prediction on this dataset.

We observe that ERFNet-HT-$\mathcal{L}_{HT}$ improves over all other models on the $1\%$ subset by a large margin. On the $1\%$ subset, there is not sufficient labelled data (less than $1K$ training images), and therefore the "ground truth" produced by pseudo-labelling in ERFNet-$pseudo$ is noisy and imperfect. 
In this case, learning from pseudo-labelled data explicitly can be harmful, while the $\mathcal{L}_{HT}$ avoids this problem by exploiting useful prior geometric knowledge about lines, in Hough space. In comparison, on the $50\%$ subset, the differences among all models are marginal, when ample training data is available. 
The experiment demonstrates the potential of our $\mathcal{L}_{HT}$ loss for data-efficient learning in Hough space in a semi-supervised setting.

We compare the performance of all ERFNet models in various driving scenarios in \tab{culane_1}.  
ERFNet-HT-$\mathcal{L}_{HT}$ shows considerable improvement over other models in most scenarios in \tab{culane_1}, and the advantage accentuates (up to $5\%$), where the amount of labelled data is decreased to $1\%$ only. 
The superiority of ERFNet-HT-$\mathcal{L}_{HT}$ demonstrates the capability of $\mathcal{L}_{HT}$ to exploit geometric lanes information from unlabelled data. 
We also notice that the "No line", "Shadow" and "Dazzle" scenarios are more challenging for all methods, compared with the other scenarios.

We visualize line predictions from different models in \fig{vis}. 
Our ERFNet-HT-$pseudo$+$\mathcal{L}_{HT}$ better localizes lanes, especially when a lane extends away from the image boundary, as in the first two examples. 
As shown in the second example, due to occlusion, ERFNet and ERFNet-HT miss the two middle lanes, while ERFNet-HT-$\mathcal{L}_{HT}$ only predicts one. 
In the third example there is an annotation inconsistency, where the opposite lane at the image border is not annotated. 
Overall, ERFNet-HT-$pseudo$+$\mathcal{L}_{HT}$ produces sharper and more precise predictions, in both simple and challenging scenarios.


%% file: sections/05_discussion.tex
\section{Limitations and conclusions}
We propose semi-supervised lane detection by exploiting global line priors in Hough space through the use of an additional loss. 
We can incorporate unlabelled data during training thus overcoming the need for expensive and error-prone annotations.
Currently our method assumes a single lane in each channel, and therefore we can optimize for the global maximum in Hough space. 
This assumption may not always hold and an extension to multiple local maxima is future research. 
However, our proposed Hough loss adds valuable prior geometric knowledge about lanes when annotations are too scarce even for pseudo-labelling based methods. 
We experimentally demonstrate the added value of our proposed loss on TuSimple and CULane datasets for limited annotated data.